\documentclass[10pt,twocolumn,letterpaper]{article}

\usepackage{cvpr}              % To produce the CAMERA-READY version
\usepackage{xspace}
\newcommand{\method}{\textsc{Newton}\xspace}

\usepackage{amsmath}
\usepackage{booktabs}
\usepackage{graphicx}
\usepackage{subcaption}
\usepackage{enumitem}
\usepackage{xcolor}
\usepackage{colortbl}

\definecolor{cvprblue}{rgb}{0.21,0.49,0.74}
\usepackage[pagebackref,breaklinks,colorlinks,allcolors=cvprblue]{hyperref}

\def\paperID{*****} % *** Enter the Paper ID here
\def\confName{CVPR}
\def\confYear{2026}

\graphicspath{{figures/}}

\title{NEWTON: Agentic Planning for Physically Grounded Video Generation}

\author{%
Yuxiang Feng$^{1,*}$\quad
Juncheng Wang$^{2,*}$\quad
Chao Xu$^{3,4}$\quad
Yijie Qian$^{1}$\quad
Huihan Wang$^{2}$\quad
Wenlong Hou$^{2}$\\
Yang Liu$^{3,4}$\quad
Baigui Sun$^{3,4}$\quad
Yong Liu$^{1,\dagger}$\quad
Shujun Wang$^{2,\dagger}$\\[0.35em]
$^{1}$Zhejiang University \quad
$^{2}$The Hong Kong Polytechnic University \\
$^{3}$IROOTECH TECHNOLOGY \quad
$^{4}$Sany Group \\[0.2em]
{\tt\small fengyx@zju.edu.cn,\ wjc2830@gmail.com,\ chaoxuxc@gmail.com,\ yijieqian@zju.edu.cn}\\
{\tt\small wang2003huihan@gmail.com,\ willen-wenlong.hou@connect.polyu.hk,\ yang.15.liu@kcl.ac.uk}\\
{\tt\small sunbaigui85@gmail.com,\ yongliu@iipc.zju.edu.cn,\ shu-jun.wang@polyu.edu.hk}\\[0.25em]
\normalsize $^{*}$Equal contribution. \quad $^{\dagger}$Corresponding authors.
}

\begin{document}
\maketitle
\begin{abstract}
Video generation models produce visually compelling results but systematically violate physical commonsense---on VideoPhy-2, the best model achieves only 32.6\% joint accuracy.
We identify a \emph{specification bottleneck}: text prompts are lossy compression of the physical world, omitting the parameters that fully determine dynamics, and no amount of model scaling can recover what was never specified.
From this diagnosis we derive three properties that physics conditioning must satisfy---sufficiency, dynamism, and verifiability---and show that no existing approach satisfies all three.
We present \method, in which video generation is demoted from the system output to one action inside an agent's toolbox: a learned planner orchestrates physics-aware tools (keyframe generation, scientific computation, prompt refinement) to construct rich conditioning, and a verifier closes the loop for iterative re-planning.
The planner is the sole trainable component, optimized on-policy via Flow-GRPO inside the live multi-turn loop.
On VideoPhy-2, \method improves joint accuracy from 21.4\% to 29.7\% on LTX-Video and from 30.7\% to 37.4\% on Veo-3.1, without modifying either generator.
Project page: \href{https://Newton026.github.io/newton}{https://Newton026.github.io/newton}.
\end{abstract}

\section{Introduction}
\label{sec:introduction}

\begin{figure}[t]
  \centering
  \includegraphics[width=.95\linewidth]{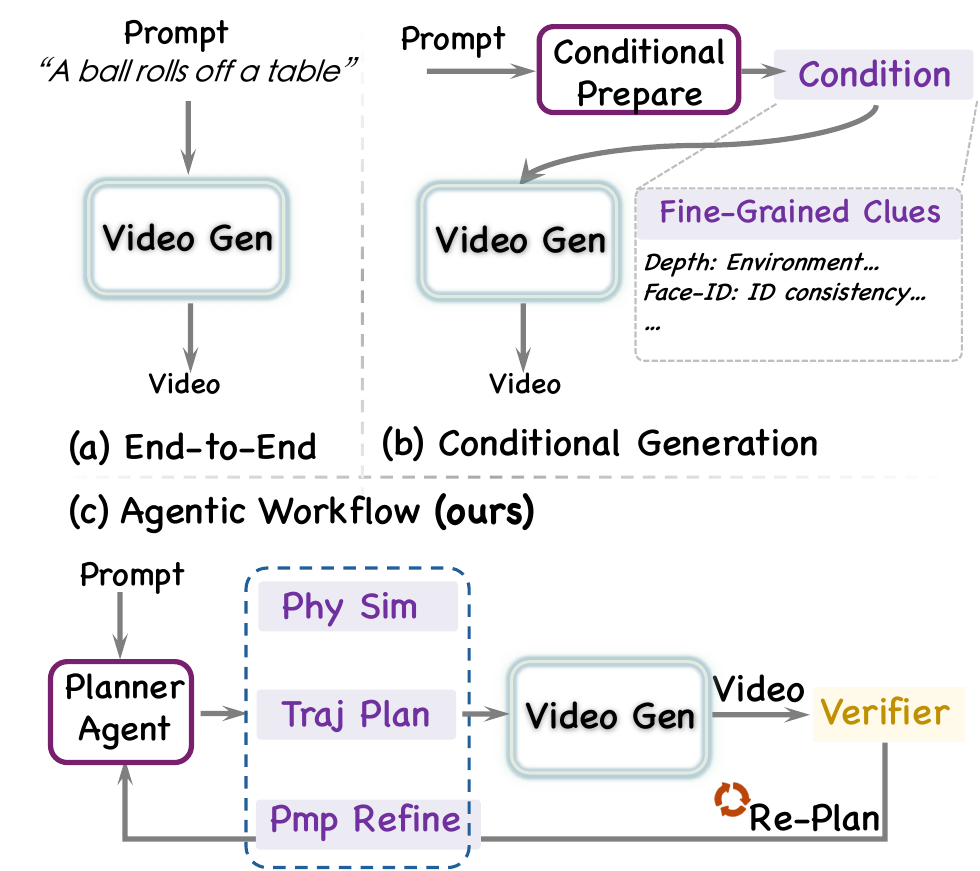}
  \caption{Three paradigms for physically grounded video generation. (a)~End-to-end: the generator hallucinates all physics from text. (b)~Conditional: fixed-modality signals (depth, identity) that cannot adapt per scene. (c)~\method (ours): a trained planner orchestrates physics tools and a verifier closes the loop for iterative re-planning.}
  \label{fig:teaser}
\end{figure}

\begin{figure*}[t]
  \centering
  \includegraphics[width=\linewidth]{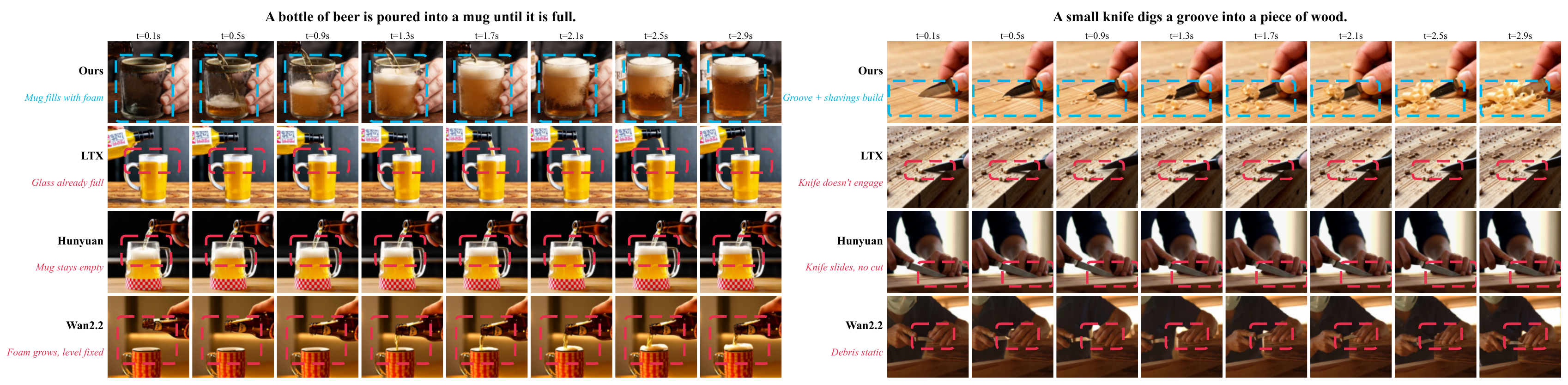}
  \caption{Left: a bottle of beer is poured into a mug until it is full---our method (top) renders progressive filling with foam buildup, while all other methods keep the liquid level fixed. Right: a small knife digs a groove into a piece of wood---our method produces a deepening groove and accumulating shavings, while none of the baselines initiate material removal. Blue boxes mark physically correct dynamics; red boxes mark violations.}
  \label{fig:vis_teaser}
\end{figure*}

Video generation has made remarkable progress.
Recent models~\cite{sora,kling,veo,wan} produce photorealistic, temporally coherent videos from text prompts, approaching the visual quality of real footage across diverse scenes and styles.

However, these models systematically fail at physics.
Balls change speed without contact, falling objects ignore gravity, and collisions violate conservation of momentum~\cite{videophy,videophy2}.
As partially shown in Fig.~\ref{fig:vis_teaser}, the failures span nearly every physical domain: Newtonian mechanics, optics, thermodynamics, and material properties~\cite{phygenbench}, as well as motion rationality and instance preservation~\cite{vbench2}.
Scaling model size or training data has not closed this gap~\cite{phygenbench,physicsiq,howfar}, pointing to a more fundamental cause.

We argue that the root cause is not insufficient capacity but insufficient \emph{specification}.
As shown by Fig.~\ref{fig:teaser}, in DiT-based generators, all guidance enters through conditioning signals---text, depth maps, motion vectors---yet text prompts are lossy compression of the physical world.
A prompt like ``a ball rolls off a table'' omits mass, friction, table height, and initial velocity, parameters that fully determine the trajectory.
The generator must hallucinate a consistent set of values from a single sentence---an ill-posed problem that produces visually plausible but physically incoherent dynamics.

From this view, we derive three properties that physics conditioning must satisfy:
(1)~\textbf{Sufficiency}---covering enough physical dimensions to determine dynamics, not leaving parameters unspecified;
(2)~\textbf{Dynamism}---adapting per scene, since different scenarios demand different physical specifications;
(3)~\textbf{Verifiability}---checking whether the output obeys the intended physics, and correcting if not.
No existing approach satisfies all three. End-to-end training embeds physics implicitly (not sufficient). ControlNet~\cite{controlnet} provides fixed-modality signals (not dynamic). All one-shot methods lack feedback (not verifiable).

Satisfying all three properties jointly requires a system that can reason about what physical knowledge a given scene demands, access heterogeneous external sources to acquire it, and iterate based on evaluation feedback.
No single-model modification can achieve this: retraining embeds physics without guarantees~\cite{howfar,phygenbench}, fixed conditioning cannot adapt across physical domains~\cite{controlnet}, and test-time search operates within the generator, unable to invoke external knowledge~\cite{videot1,phyt2v}.
These capabilities---adaptive reasoning, heterogeneous tool use, and closed-loop correction---are precisely what characterizes an autonomous agent, which raises a natural question: \emph{how can we build an agentic system that reasons about missing physics per scene, acquires it through external tools, and iteratively corrects generation---all without modifying the generator itself?}

We present \method (\textbf{Ne}ural Agentic \textbf{W}orld-Aware \textbf{T}ool-\textbf{O}rchestrated \textbf{N}avigation), in which video generation is demoted from the system output to one action inside an agent's toolbox.
It consists of three components: a \emph{Planner} that decides which physics-aware tools to invoke for a given prompt, an \emph{Executor} that dispatches those tools alongside the frozen video generator, and a \emph{Verifier} that scores the resulting video on physical plausibility.
These components operate in an iterative loop: at each cycle, the Planner reads prior feedback and selects tools to construct richer conditioning, the Executor produces a video, and the Verifier evaluates it---feeding scores back for re-planning.
\textbf{Only the Planner is trainable}; it is optimized on-policy via Flow-GRPO~\cite{agentflow} inside the live multi-turn loop, while the tool library, the video generator, and the Verifier all remain frozen.
This architecture directly maps onto the three requirements: the \emph{tool library} provides \textbf{sufficiency} by covering complementary physical dimensions; the \emph{Planner} provides \textbf{dynamism}, selecting and composing tools per scene; and the \emph{verify--correct loop} provides \textbf{verifiability}, feeding evaluation back for re-planning.

\method substantially improves physical commonsense on two frozen generators (LTX-Video and Veo-3.1) without modifying them.
The planner learns scene-dependent tool scheduling---computing trajectories for projectiles, generating keyframes for spatial constraints, refining prompts for material properties.
Physical consistency shifts from hoping for emergence to engineering it through agentic planning.

In summary, our contributions are:
\begin{itemize}[leftmargin=*, nosep]
  \item We identify the \emph{specification bottleneck} as the root cause of physics failures in video generation, and derive three necessary properties---sufficiency, dynamism, and verifiability---that any physics conditioning must satisfy.
  \item We propose \method, an agentic framework that demotes video generation from the system output to one action in a planner's toolbox, orchestrating physics-aware tools and a verifier in an iterative loop.
  \item We introduce a training recipe in which the planner---the sole trainable component---is optimized on-policy via Flow-GRPO inside the live multi-turn loop, requiring no modification to the frozen video generator.
  \item We demonstrate substantial improvements on VideoPhy-2 across two generators, showing that the planner discovers scene-dependent tool-use strategies that generalize across unseen physical scenarios.
\end{itemize}

\section{Related Work}
\label{sec:related}

\subsection{Physics-Grounded Video Generation}
Video generation has received tremendous attention in recent years. Closed systems such as Sora~\cite{sora}, Veo~\cite{veo} and Kling~\cite{kling}, together with open-weight models Wan~\cite{wan}, LTX-Video~\cite{ltx}, Hunyuan-Video~\cite{ltx}, produce photorealistic clips with strong text adherence and camera control. Despite rapid scaling, this surface is fundamentally underspecified for dynamics, and a growing body of physics-grounded video generation~\cite{physanimator, phantom, phystalk, phyco} has emerged to close the gap.

One line of work treats an explicit simulator as a prior. PhysMotion~\cite{physmotion} time-steps a coarse 3D Gaussian object with differentiable MPM and refines frames with a T2I model. PhysCtrl~\cite{physctrl} trains a generative physics network over 550K simulated trajectories spanning four materials (elastic, sand, plasticine, rigid). PhysChoreo~\cite{physchoreo} further introduces part-aware material-field reconstruction from a single image and drives a generator with a temporally instructed, physically editable simulator. These methods deliver strong continuum-mechanics behavior but commit to a fixed simulator family and do not adapt the tooling to the scene. Rather than calling an external simulator, NewtonGen~\cite{newtongen} embeds Neural Newtonian Dynamics linear physics-informed Neural ODEs with a residual MLP. The formulation is elegant for single-object continuous motion but, by construction, struggles with collisions and multi-object interaction.

A complementary direction modifies the generator itself to internalize physics. VideoREPA distills token-level relations from a self-supervised video foundation model into a DiT, narrowing a measurable physics-understanding gap on Physion. WISA~\cite{wisa} decomposes physics into hierarchical textual, qualitative, and quantitative signals injected through a Mixture-of-Physical-Experts attention block paired with the WISA-80K dataset. ProPhy~\cite{prophy} pushes this further with a two-stage Mixture-of-Physics-Experts and a VLM-distilled refinement block that produces anisotropic, region-level physical alignment. Reward-based post-training such as PhyGDPO~\cite{phygdpo} shifts the implicit prior in a similar one-shot manner, without per-sample verification.

Across these directions, no method jointly satisfies the sufficiency, dynamism and verifiability properties identified in the Introduction, which are addressed in \method.

\subsection{Agentic Systems for Visual Generation}
We follow the line of agentic LLM systems in which a planner decomposes a high-level goal, selects from an external tool library, executes the chosen tool, and critiques the result before re-planning~\cite{react, toolformer}. Recent works~\cite{agentic,gtpo,landscape} have emphasized that the agent itself, not only its tools, benefits from being trainable on-policy rather than driven by a frozen prompted LLM. For example, AgentFlow~\cite{agentflow} demonstrated that a planner--executor--verifier--generator stack with on-policy Flow-GRPO~\cite{flowgrpo} training can substantially outperform frozen orchestration on text reasoning tasks.

This framing has been productive in image generation. GenAgent~\cite{genagent} decouples understanding and generation by treating image generators as invokable tools, then trains the agent end-to-end with agentic RL combining pointwise quality and pairwise reflection rewards. M3~\cite{m3} orchestrates a Planner--Checker--Refiner--Editor--Verifier ensemble that iteratively repairs compositional failures at inference time. coDrawAgents~\cite{codrawagents} runs an Interpreter--Planner--Checker--Painter dialogue with explicit error correction over layouts before rendering.

Agentic ideas have only recently reached video generation~\cite{agenticvideo,moregen}. Closest to us is the Chain of Event-Centric Causal Thought (CECT) framework~\cite{cect}, which uses an LLM to reason about a sequence of physically plausible events and conditions a video diffusion model on this causal chain, directly attacking the failure mode that diffusion renders physics as a single moment rather than a causal progression. Our setting differs from CECT in three respects. (i)~\emph{Tools, not text.} CECT outputs an enriched textual event chain; \method wields a heterogeneous tool library---keyframe generation, Python physical computation, prompt refinement---whose outputs are explicit physical signals that a prompt alone cannot carry. (ii)~\emph{Verification in the loop.} CECT plans once; \method closes a verify--correct loop via VideoPhy-2-AutoEval~\cite{videophy2} and re-plans for up to five iterations per scene. (iii)~\emph{On-policy planning.} Where CECT relies on the frozen reasoning of a generic LLM, our planner is trained on-policy with Flow-GRPO inside the live loop, so it learns which tool to invoke when against the realized verifier signal. Together these distinctions move physical reasoning from prompt engineering to engineered, agentic control.

\section{Preliminary and Motivation}
\label{sec:prelim}

\subsection{Video Generation with Diffusion Transformers}
\label{sec:dit}

Modern text-to-video generators build on the Diffusion Transformer (DiT) architecture.
A pretrained VAE encodes a video $\mathbf{x} \in \mathbb{R}^{F \times H \times W \times 3}$ into a latent $\mathbf{z} \in \mathbb{R}^{f \times h \times w \times d}$, which is patchified into tokens and processed by transformer blocks.
The model is trained via flow matching: given an interpolation $\mathbf{z}_t = (1{-}t)\,\boldsymbol{\epsilon} + t\,\mathbf{z}$ between noise $\boldsymbol{\epsilon}$ and clean latent $\mathbf{z}$, it learns a velocity field $\mathbf{u}_\theta$ by minimizing
\begin{equation}
  \mathcal{L}_{\mathrm{flow}} = \mathbb{E}_{t,\, \mathbf{z},\, \boldsymbol{\epsilon}} \big\lVert \mathbf{u}_\theta(\mathbf{z}_t, t;\, C) - (\mathbf{z} - \boldsymbol{\epsilon}) \big\rVert_2^2,
  \label{eq:flow}
\end{equation}
where $C$ is the conditioning context.
At inference, an ODE solver integrates from noise ($t{=}0$) to data ($t{=}1$).

The conditioning interface $C$ accepts heterogeneous signals---text tokens from language encoders and image tokens from visual encoders---via cross-attention or adaptive normalization.
This multi-modal interface means the generator can be steered by both text prompts and reference images without architectural change.
A direct consequence: \emph{generation quality is bounded by conditioning quality}.

\begin{figure}[t]
  \centering
  \includegraphics[width=\linewidth]{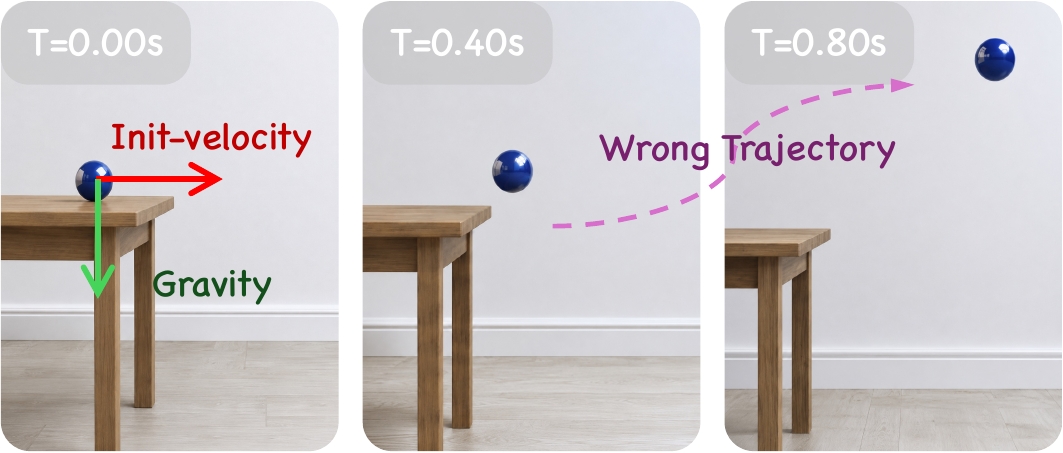}
  \caption{Text prompts are a lossy compression of physics: the same caption is consistent with many physically distinct trajectories, none of which the generator can disambiguate without additional conditioning.}
  \label{fig:motivation}
\end{figure}

\subsection{Motivation: The Specification Bottleneck}
\label{sec:motivation}

Despite strong visual fidelity, current generators systematically violate physical commonsense.
On VideoPhy-2~\cite{videophy2}, even the best model achieves only 32.6\% joint performance (videos with both SA$\geq$4 and PC$\geq$4), with conservation-law violations approaching 40\%.

\paragraph{Text prompts are lossy compression of the physical world.}
The root cause is insufficient specification, not insufficient capacity.
Consider ``a ball rolls off the edge of a table''---this sentence omits the ball's mass, the friction coefficient, the table height, the initial velocity, and the surface material below, all of which jointly determine the physical trajectory.
As shown in Fig.~\ref{fig:motivation}, the generator must hallucinate a consistent set of these parameters from a single sentence---an ill-posed problem that produces visually plausible but physically incoherent dynamics.

\paragraph{Human physics knowledge remains untapped.}
Humans have spent millennia building structured physical laws---Newtonian mechanics, conservation principles, fluid dynamics---that can fully determine trajectories given the relevant parameters.
Current generators instead learn physics implicitly from raw video, akin to rediscovering Newton's laws from unlabeled footage.
This is both data-inefficient and fundamentally limited by training coverage.

\paragraph{From rendering to specification.}
These observations suggest a different strategy: rather than retraining the generator, \emph{enrich its conditioning signal} with physics knowledge.
If we provide physically grounded keyframes, quantitative constraints, and precise prompts, the generator's existing capacity suffices to render plausible physics.
The remaining challenge---automatically acquiring and structuring the right physical knowledge for a given prompt---motivates \method.

\section{\method: Neural Agentic World-Aware Tool-Orchestrated Navigation}
\label{sec:method}

\begin{figure*}[t]
  \centering
  \includegraphics[width=0.95\linewidth]{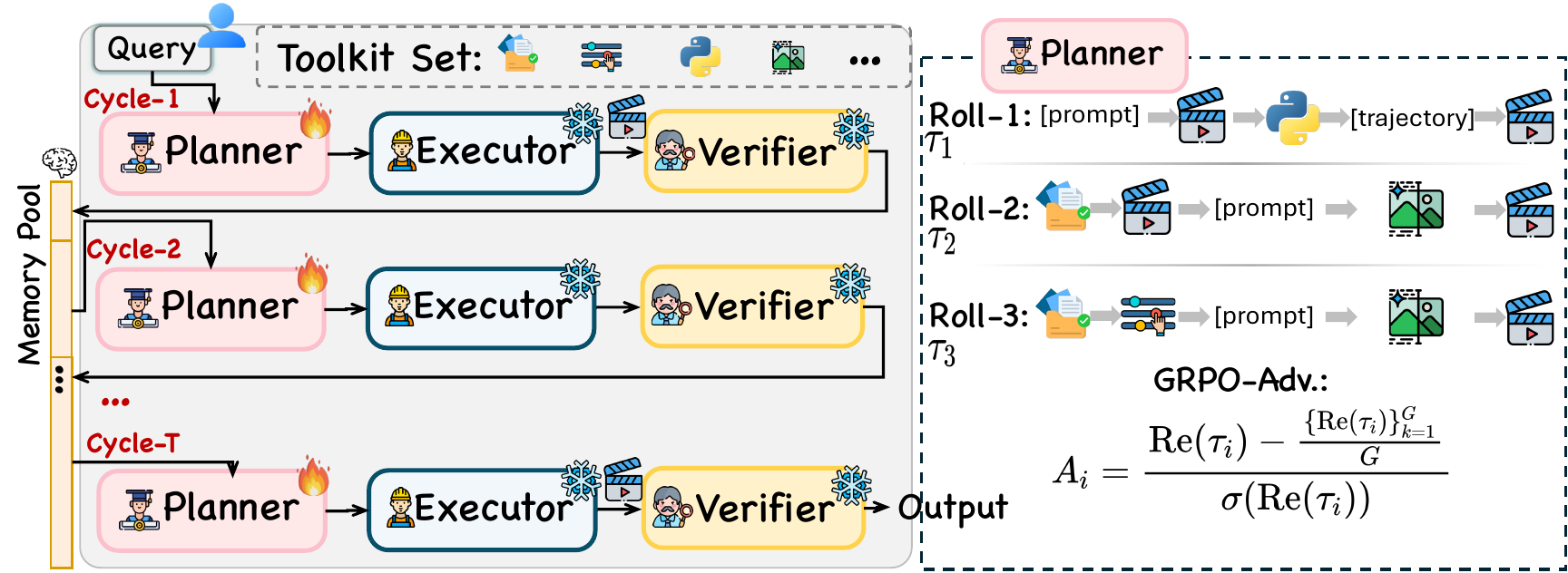}
  \caption{\method overview. \textbf{Left}: the iterative pipeline. A user query and toolkit set initialize Cycle-1; at each cycle the Planner (trainable) reads the memory pool, selects tools, and the Executor dispatches them alongside the frozen video generator. The Verifier (frozen) scores the result on SA and PC, appending feedback to memory for the next cycle. The best-scored video across $T$ cycles is returned. \textbf{Right}: Flow-GRPO training. $G$ parallel rollouts $\{\tau_i\}$ are sampled under the current policy; each executes the full $T$-cycle trajectory. The group-normalized advantage $A_i$ drives a clipped surrogate update on the Planner alone.}
  \label{fig:system}
\end{figure*}

\method is a trainable agentic system that improves the physical plausibility of videos from a frozen generator by enriching its conditioning signal with physics knowledge.
It consists of three components: a \emph{Planner} that decides which physics-aware tools to invoke for a given prompt, an \emph{Executor} that dispatches those tools alongside the frozen video generator, and a \emph{Verifier} that scores the resulting video on physical plausibility.
These components operate in an iterative loop: at each cycle, the Planner reads prior feedback and selects tools to construct richer conditioning, the Executor produces a video, and the Verifier evaluates it---feeding scores back for re-planning.
\textbf{Only the Planner is trainable}; it is optimized on-policy via Flow-GRPO~\cite{agentflow} inside this live multi-turn loop, while the tool library, the video generator, and the Verifier all remain frozen.

\subsection{System Pipeline}
\label{sec:pipeline}

\subsubsection{Three-Role Architecture}

Inspired by prior work on agentic planning and verification~\cite{innermonologue,react,reflexion,agentflow}, \method decomposes video generation into three roles, shown in Fig.~\ref{fig:system}.

\paragraph{Planner.}
A vision--language model (VLM) serves as the sole trainable component.
At each cycle $t$, it reads the memory state $M^t$---original prompt, prior tool calls and outputs, verifier feedback---and produces a structured action $a^t \sim \pi_\theta(a^t \mid q, M^t)$ specifying which tools to invoke and with what arguments.
The action space is flexible: the Planner may call any subset of tools, trigger video generation, or skip a cycle entirely.

\paragraph{Executor.}
The Executor carries out the Planner's actions by dispatching calls to three physics-aware tools (\S\ref{sec:tools_detail}) and the frozen video generator.
When video generation is triggered, the generator is conditioned on accumulated tool outputs---refined prompts as text and keyframes as images---with the specific mechanism depending on the generator's interface.
The framework is \textbf{generator-agnostic}.

\paragraph{Verifier.}
A multimodal evaluation model rates each generated video on two scalar dimensions: Semantic Adherence (SA) and Physical Commonsense (PC).
Scores are appended to memory, closing the feedback loop.

\subsubsection{Iterative Cycle}

The system runs for $T$ fixed cycles, formalized as a finite-horizon MDP.
At cycle $t$, the Planner observes $M^t$, selects action $a^t$, and the Executor produces observation $e^t$.
The memory updates deterministically: $M^{t+1} = f_{\mathrm{mem}}(M^t, a^t, e^t)$.
Not every cycle must produce a video---early cycles may focus on computation and prompt refinement, while later cycles leverage accumulated knowledge for generation.
The video with the highest verifier score across all cycles is returned as the final output.

The memory $M^t$ stores all prior context---Planner reasoning, tool arguments and outputs, verifier scores---but excludes generated videos to keep context length tractable; the verifier's scalar scores serve as a sufficient summary.

\subsubsection{Physics-Aware Tools}
\label{sec:tools_detail}

Three tools target complementary dimensions of the specification bottleneck.

\paragraph{Keyframe Generation.}
A text-to-image model generates guiding images at designated temporal positions (e.g., first, middle, and last frames).
The Planner writes a dedicated prompt for each keyframe encoding the expected physical state (e.g., ``ball at the apex of a parabolic arc'' for the mid-frame).
These keyframes impose \textbf{temporal boundary conditions}, anchoring the trajectory at physically consistent states and constraining the generator's interpolation.

\paragraph{Python Computation.}
Provides a sandboxed Python environment for scientific computation---projectile trajectories, conservation-of-momentum calculations, rotational dynamics.
Numerical results enter memory and inform subsequent keyframe prompts or constraint specification, operationalizing the human physics knowledge identified in \S\ref{sec:motivation}.

\paragraph{Prompt Refiner.}
Performs natural-language refinement of the generation prompt, augmenting it with physical detail, material properties, or scene constraints absent from the original caption.

\subsection{In-the-Flow Optimization}
\label{sec:optimization}

\subsubsection{Why In-the-Flow}

Offline supervised training decouples the Planner from live system dynamics: it never observes its own mistakes, cannot recover from tool failures, and does not adapt to actual verifier feedback.
AgentFlow~\cite{agentflow} shows that SFT on expert trajectories causes a 19\% average accuracy drop versus a frozen baseline in agentic settings.
We instead train the Planner \textbf{in the flow of execution}, rolling out the full system under the current policy and updating based on actual outcomes.

\subsubsection{Flow-GRPO}

We adopt Flow-GRPO~\cite{agentflow}, an on-policy algorithm for multi-turn agents with sparse rewards.
It broadcasts a single trajectory-level reward to every cycle, converting multi-turn credit assignment into tractable single-turn updates.

For each prompt $q$, we sample $G$ parallel rollouts $\{\tau_i\}_{i=1}^G$ under $\pi_{\theta_{\mathrm{old}}}$, where each rollout executes the full $T$-cycle trajectory $\tau_i = \{(a_i^t, e_i^t)\}_{t=1}^{T}$---the Planner makes all $T$ decisions before a reward is assigned, ensuring the policy is exposed to the complete planning horizon.
The per-rollout advantage is group-normalized:
\begin{equation}
  A_i = \frac{R(\tau_i) - \mathrm{mean}(\{R(\tau_k)\}_{k=1}^{G})}{\mathrm{std}(\{R(\tau_k)\}_{k=1}^{G})}.
  \label{eq:advantage}
\end{equation}
The policy is updated via the clipped surrogate objective:
\begin{equation}
\begin{aligned}
  \mathcal{J}(\theta) &= \mathbb{E}\bigg[\frac{1}{G}\sum_{i=1}^{G} \frac{1}{T} \sum_{t=1}^{T} \frac{1}{|a_i^t|} \sum_{j=1}^{|a_i^t|} \\
  &\quad\; \min\!\big\{\rho_{i,j}^{t}\, A_i,\; \mathrm{clip}(\rho_{i,j}^{t}, 1{-}\epsilon, 1{+}\epsilon)\, A_i\big\}\bigg] \\
  &\quad - \beta\, \mathbb{D}_{\mathrm{KL}}(\pi_\theta \| \pi_{\mathrm{ref}}),
\end{aligned}
  \label{eq:flow_grpo}
\end{equation}
where $\rho_{i,j}^{t}$ is the token-level importance ratio, $\epsilon$ the clipping parameter, and $\beta$ the KL penalty weight against a fixed reference policy $\pi_{\mathrm{ref}}$.

\subsubsection{Reward Design}

The composite reward has three components:
\begin{equation}
  R(\tau) = R_{\mathrm{quality}} + R_{\mathrm{kf}} + R_{\mathrm{compute}}.
  \label{eq:reward}
\end{equation}

\paragraph{Format penalty.}
Any format or length violation in any cycle triggers a fixed negative reward, enforcing the basic interface contract.

\paragraph{Quality reward $R_{\mathrm{quality}}$.}
A tiered function of the maximum SA and PC scores across all video-producing cycles.
Rather than a binary pass/fail, we introduce intermediate tiers that reward partial physical correctness (e.g., high SA with moderate PC, or vice versa), densifying the advantage signal in a domain where joint high scores are rare.

\paragraph{Keyframe bonus $R_{\mathrm{kf}}$.}
A fixed bonus awarded when a cycle uses newly generated keyframes for conditioning and the resulting video meets a semantic-adherence threshold.
This term is independent of $R_{\mathrm{quality}}$, encouraging keyframe exploration early in training.

\paragraph{Computation bonus $R_{\mathrm{compute}}$.}
A fixed bonus awarded when the trajectory contains a valid physics computation (correct function and parameters) \emph{and} the quality reward is positive.
The conjunction prevents reward hacking from vacuous computations.

The tiered quality reward and independent tool-use bonuses together yield a dense set of achievable reward values, enabling effective group-normalized advantage estimation.

\section{Experiments}
\label{sec:exp}

\begin{figure*}[!tp]
    \centering
    \includegraphics[width=\linewidth]{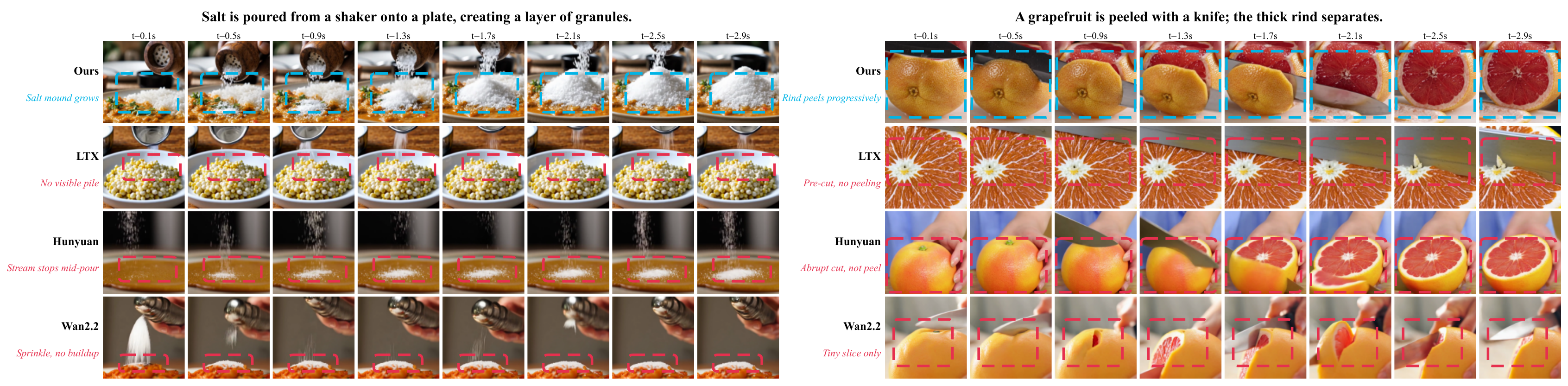}
    \caption{Qualitative comparison on real-world samples.}
    \label{fig:vis_real}
\end{figure*}

\begin{figure*}[!tp]
    \centering
    \includegraphics[width=\linewidth]{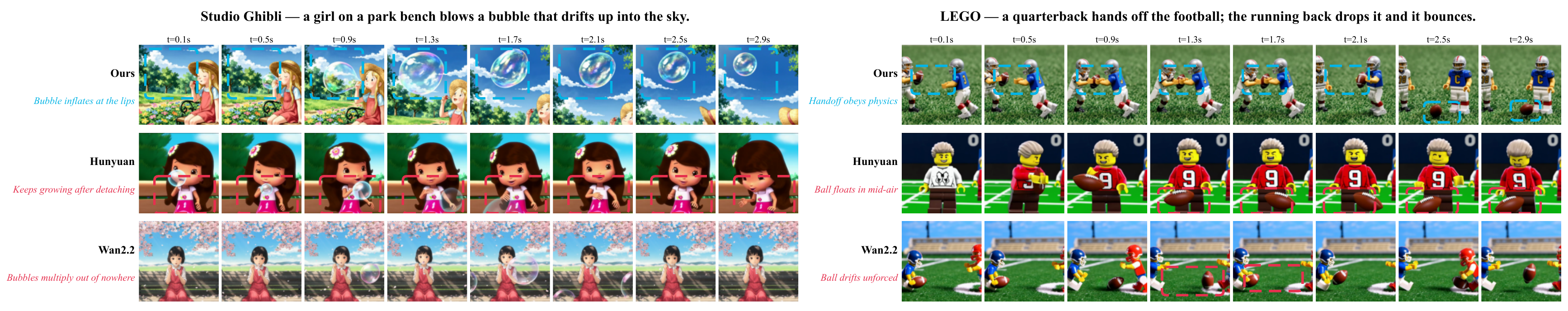}
    \caption{Qualitative comparison on animation samples.}
    \label{fig:vis_animation}
\end{figure*}

\begin{figure}[!tp]
  \centering
  \includegraphics[width=0.85\linewidth]{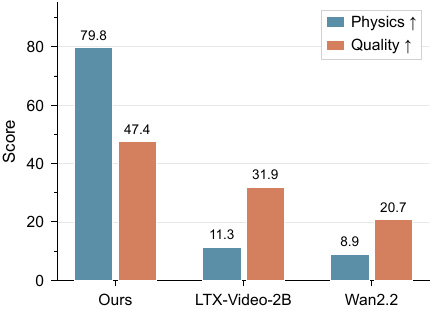}
  \caption{Results of human preference study.}
  \label{fig:human}
\end{figure}

We evaluate \method on a primary physics benchmark (\S\ref{sec:exp_main}), a held-out cross-benchmark (\S\ref{sec:exp_phygen}), and four ablations on the design axes \method introduces (\S\ref{sec:exp_ablation}).

\subsection{Experimental Setup}
\label{sec:exp_setup}

\paragraph{Benchmarks.}
\textbf{VideoPhy-2}~\cite{videophy2} is our primary benchmark: 590 captions across 197 physical actions, with a designated \textsc{Hard} subset of 180 captions targeting conservation laws, multi-object collisions, and articulated dynamics.
Each video is rated on Semantic Adherence (SA) and Physical Commonsense (PC); we report the percentage passing $\mathrm{PC}{\geq}4$ (PC), $\mathrm{SA}{\geq}4$ (SA), and both jointly (Joint).
\textbf{PhyGenBench}~\cite{phygenbench} provides 160 prompts across Mechanics, Optics, Thermal, and Material, scored by its official VLM-judged protocol on $[0,1]$.

\paragraph{Baselines.}
(i)~\emph{Strong text-to-video generators}: Wan2.2-TI2V-5B~\cite{wan}, Cosmos-Predict2.5~\cite{cosmos}, HunyuanVideo~\cite{hunyuanvideo}, CogVideoX-5B~\cite{cogvideox}, and LTX-Video-2B~\cite{ltx}.
(ii)~\emph{Physics-augmented generators} on top of LTX-Video-2B: VideoREPA~\cite{videorepa} and WISA~\cite{wisa}, both re-implemented on LTX-Video-2B.

\paragraph{Implementation.}
The Planner (Qwen3.5-9B) is the only trainable module; the video generator (LTX-Video-2B unless stated) and the VideoPhy-2-AutoEval verifier stay frozen.
We optimize with Flow-GRPO on the 3{,}350-prompt VideoPhy-2 train split for one epoch: $G{=}8$ rollouts, $T{=}5$ cycles, $\epsilon{=}0.2$, $\beta{=}0.01$, entropy coefficient $0.005$, learning rate $5{\times}10^{-7}$, training batch $4$ / PPO mini-batch $32$ / per-GPU micro-batch $8$, on $8$ NVIDIA H200 GPUs.

\subsection{Main Results on VideoPhy-2}
\label{sec:exp_main}

\begin{table}[!ht]
  \caption{Physical plausibility on VideoPhy-2 (\%). Column groups: full 590-caption set (\textsc{All}) and the 180-caption \textsc{Hard} subset. PC = Physical Commonsense, SA = Semantic Adherence, Joint = both $\geq 4$. Subscript $\textsc{H}$ denotes the \textsc{Hard} subset. \colorbox{gray!18}{Best} per column. \textsc{+Ours} attaches \method to the same LTX-Video-2B backbone used by VideoREPA and WISA.}
  \label{tab:main}
  \scriptsize
  \setlength{\tabcolsep}{2.6pt}
  \begin{tabular}{lcccccc}
    \toprule
    Method & PC & SA & Joint & PC$_{\textsc{H}}$ & SA$_{\textsc{H}}$ & Joint$_{\textsc{H}}$ \\
    \midrule
    Wan2.2-TI2V-5B~\cite{wan}            & 58.00 & 28.30 & 24.20 & 37.20 &  8.30 &  3.90 \\
    Cosmos-Predict2.5~\cite{cosmos}      & 58.64 & 28.31 & 22.71 & 46.11 & 12.22 &  7.78 \\
    HunyuanVideo~\cite{hunyuanvideo}          & 73.56 & 25.25 & 22.54 & 63.33 &  8.89 &  6.67 \\
    CogVideoX-5B~\cite{cogvideox}        & 69.15 & 30.17 & 25.93 & 54.44 &  9.44 &  5.00 \\
    LTX-Video-2B~\cite{ltx}              & 74.25 & 22.88 & 21.36 & 63.33 &  6.67 &  4.44 \\
    \midrule
    \;+ VideoREPA~\cite{videorepa}       & \colorbox{gray!18}{84.40} &  6.60 &  5.30 & \colorbox{gray!18}{86.10} &  1.10 &  1.10 \\
    \;+ WISA~\cite{wisa}                 & 77.30 & 10.80 &  9.80 & 74.40 &  2.20 &  2.20 \\
    \;+ \method (Ours)                   & 82.71 & \colorbox{gray!18}{31.53} & \colorbox{gray!18}{29.66} & 78.89 & \colorbox{gray!18}{12.78} & \colorbox{gray!18}{12.22} \\
    \bottomrule
  \end{tabular}
\end{table}

Table~\ref{tab:main} reports VideoPhy-2.
\method is the only method that improves \emph{both} PC and SA over its LTX-Video-2B backbone, lifting Joint accuracy from 21.36\% to 29.66\% on the full set and from 4.44\% to 12.22\% on \textsc{Hard} (a $2.75{\times}$ relative gain).
VideoREPA and WISA show a sharp PC--SA trade-off: VideoREPA tops PC (84.4\% / 86.1\%) but its SA collapses below 7\%, dragging Joint \emph{below} the LTX-Video baseline; WISA exhibits the same pattern at smaller magnitude.

We also provide two qualitative comparisons in Fig.~\ref{fig:vis_real}. Left: salt pouring---\method shows the salt mound progressively building on the plate, while LTX-Video produces no visible pile, Hunyuan stops the stream mid-pour, and Wan2.2 sprinkles without accumulation. Right: grapefruit peeling---\method renders the rind progressively separating from the flesh, while baselines either start pre-cut, perform an abrupt cut without peeling, or produce only a tiny slice. Fig.~\ref{fig:vis_animation} shows qualitative comparison on animation samples.

Furthermore, we conduct a controlled human preference study by sampling one prompt per physical action from VideoPhy-2, yielding 197 prompts, and generating three videos per prompt with LTX-Video-2B, Wan2.2, and \method. The three videos are displayed simultaneously in randomized horizontal order with method identities hidden. We recruited 20 volunteers, each of whom answered two independent forced-choice questions per triplet---which video best obeys the implied physics? and which video has the best overall quality? The results in Fig.~\ref{fig:human} indicate that our model consistently outperforms baseline methods in terms of both physics plausibility and video quality.

\subsection{Cross-Benchmark Generalization on PhyGenBench}
\label{sec:exp_phygen}

\begin{table}[!ht]
  \caption{PhyGenBench results (VLM-judged score, $\uparrow$). \method generalizes from its VideoPhy-2 training distribution to the four physical categories of PhyGenBench. \colorbox{gray!18}{Best} per column.}
  \label{tab:phygen}
  \small
  \setlength{\tabcolsep}{4.2pt}
  \begin{tabular}{lccccc}
    \toprule
    Method & Mech. & Optics & Therm. & Mat. & Avg. \\
    \midrule
    Wan2.2-TI2V-5B    & \colorbox{gray!18}{0.550} & 0.580 & \colorbox{gray!18}{0.533} & 0.500 & 0.544 \\
    Cosmos-Predict2.5 & 0.300 & 0.333 & 0.411 & 0.385 & 0.352 \\
    HunyuanVideo      & 0.325 & 0.387 & 0.256 & 0.299 & 0.325 \\
    CogVideoX-5B      & 0.325 & 0.360 & 0.378 & 0.359 & 0.354 \\
    LTX-Video-2B      & 0.508 & 0.580 & 0.478 & 0.450 & 0.510 \\
    \;+ \method       & 0.500 & \colorbox{gray!18}{0.647} & 0.522 & \colorbox{gray!18}{0.542} & \colorbox{gray!18}{0.560} \\
    \bottomrule
  \end{tabular}
\end{table}

Table~\ref{tab:phygen} evaluates the same trained planner---without retraining---on PhyGenBench.
\method raises the average from 0.510 to 0.560, surpassing the previously strongest open generator (Wan2.2-TI2V-5B at 0.544) with a $2.5{\times}$ smaller backbone.
Gains concentrate on Optics ($+0.067$) and Material ($+0.092$); Mechanics is essentially unchanged ($-0.008$).

\subsection{Ablation Studies}
\label{sec:exp_ablation}

\paragraph{Planner scale.}
Table~\ref{tab:abl_size} sweeps the Qwen3.5 planner across 2B / 4B / 9B.
\textsc{Hard}-Joint rises monotonically (7.22\% $\to$ 9.44\% $\to$ 12.22\%), and the 9B planner consistently leads on every column.

\begin{table}[!ht]
  \centering
  \footnotesize
  \setlength{\tabcolsep}{3pt}
  \caption{\textbf{Planner scale.} Larger planners help most on \textsc{Hard}. Subscript $\textsc{H}$ denotes the \textsc{Hard} subset. \colorbox{gray!18}{Best} per column.}
  \label{tab:abl_size}
  \begin{tabular}{@{}l cccccc@{}}
    \toprule
    Planner & PC & SA & Joint & PC$_{\textsc{H}}$ & SA$_{\textsc{H}}$ & Joint$_{\textsc{H}}$ \\
    \midrule
    Qwen3.5-2B & 82.20 & 27.63 & 25.76 & 74.44 & 8.89 & 7.22 \\
    Qwen3.5-4B & 81.36 & 28.81 & 26.44 & 72.22 & 8.33 & 5.00 \\
    Qwen3.5-9B & \colorbox{gray!18}{82.71} & \colorbox{gray!18}{31.53} & \colorbox{gray!18}{29.66} & \colorbox{gray!18}{78.89} & \colorbox{gray!18}{12.78} & \colorbox{gray!18}{12.22} \\
    \bottomrule
  \end{tabular}
\end{table}

\paragraph{Number of planning cycles.}
Table~\ref{tab:abl_turns} sweeps $T \in \{2, 3, 5\}$, training a separate planner at each $T$ and evaluating under the matching budget.
\textsc{Hard}-Joint climbs from 4.44\% ($T{=}2$) to 10.00\% ($T{=}3$) to 12.22\% ($T{=}5$).

\begin{table}[!ht]
  \centering
  \footnotesize
  \setlength{\tabcolsep}{3pt}
  \caption{\textbf{Cycle budget $T$.} A separate planner is trained for each $T$ and evaluated under the matching cycle budget. Verify-and-correct gains compound up to $T{=}5$. \colorbox{gray!18}{Best} per column.}
  \label{tab:abl_turns}
  \begin{tabular}{@{}c cccccc@{}}
    \toprule
    $T$ & PC & SA & Joint & PC$_{\textsc{H}}$ & SA$_{\textsc{H}}$ & Joint$_{\textsc{H}}$ \\
    \midrule
    2 & 79.32 & 29.49 & 27.80 & 69.44 & 10.00 & 4.44 \\
    3 & 82.88 & 31.02 & 28.98 & 75.00 & 12.22 & 10.00 \\
    5 & \colorbox{gray!18}{82.71} & \colorbox{gray!18}{31.53} & \colorbox{gray!18}{29.66} & \colorbox{gray!18}{78.89} & \colorbox{gray!18}{12.78} & \colorbox{gray!18}{12.22} \\
    \bottomrule
  \end{tabular}
\end{table}

\begin{figure}[!ht]
  \centering
  \includegraphics[width=\linewidth]{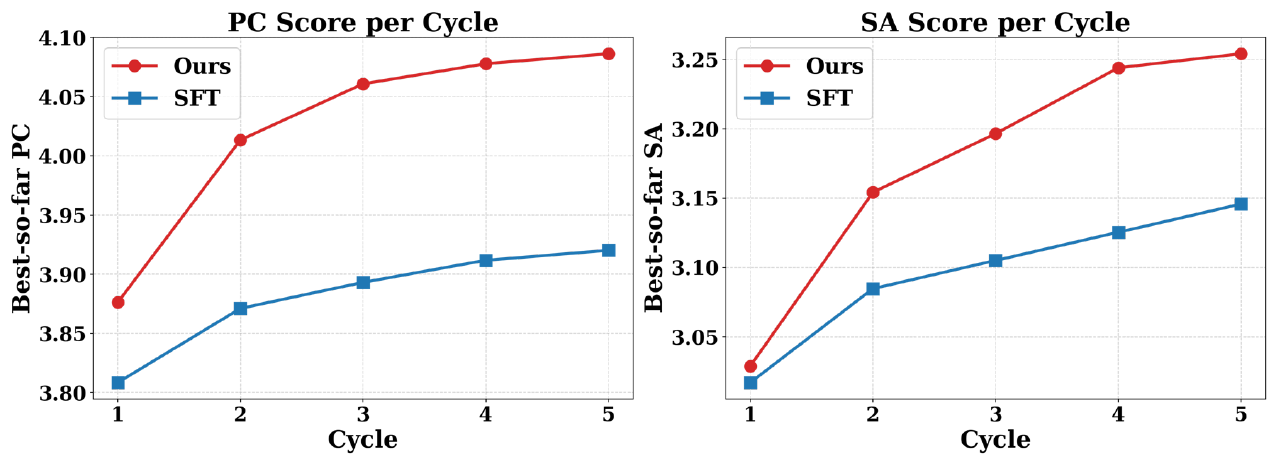}
  \caption{Best-so-far PC and SA on VideoPhy-2 (590 prompts) across the five refinement cycles. \textsc{Ours} widens the gap to offline SFT with each cycle.}
  \label{fig:bestsofar}
\end{figure}

\paragraph{Training strategy.}
Table~\ref{tab:abl_train} compares three regimes for the same Qwen3.5-9B Planner: \textsc{Frozen} (prompted only), \textsc{Offline SFT} (on high-reward GPT-5.4-mini rollouts collected on the train split and filtered by verifier score), and \textsc{Flow-GRPO (Ours)} (on-policy inside the live multi-turn loop).
SFT improves modestly over \textsc{Frozen}, and Flow-GRPO roughly \emph{doubles} every one of those gains (e.g., \textsc{Hard}-Joint $+1.7$ vs.\ $+3.3$; PC $+2.7$ vs.\ $+5.9$).
Fig.~\ref{fig:bestsofar} shows the same effect per cycle: \textsc{Ours} gains $\sim 2{\times}$ as much as SFT each refinement step (PC $+0.21$ vs.\ $+0.11$; SA $+0.22$ vs.\ $+0.13$ across cycles 1--5).

\begin{table}[!ht]
  \centering
  \footnotesize
  \setlength{\tabcolsep}{3pt}
  \caption{\textbf{Planner training strategy.} On-policy Flow-GRPO clearly leads, roughly doubling the gain that offline \textsc{SFT} obtains over the \textsc{Frozen} baseline. \colorbox{gray!18}{Best} per column.}
  \label{tab:abl_train}
  \begin{tabular}{@{}l cccccc@{}}
    \toprule
    Training & PC & SA & Joint & PC$_{\textsc{H}}$ & SA$_{\textsc{H}}$ & Joint$_{\textsc{H}}$ \\
    \midrule
    Frozen           & 76.78 & 31.02 & 28.14 & 67.78 & 11.67 & 8.89 \\
    Offline SFT      & 79.49 & 31.36 & 28.81 & 72.78 & 12.22 & 10.56 \\
    Flow-GRPO (Ours) & \colorbox{gray!18}{82.71} & \colorbox{gray!18}{31.53} & \colorbox{gray!18}{29.66} & \colorbox{gray!18}{78.89} & \colorbox{gray!18}{12.78} & \colorbox{gray!18}{12.22} \\
    \bottomrule
  \end{tabular}
\end{table}

\begin{table}[!htbp]
  \centering
  \footnotesize
  \setlength{\tabcolsep}{4pt}
  \caption{\textbf{Generator backbone.} Attaching \method to two different frozen generators on a 100-caption subset of the VideoPhy-2 test set (sub-sampled from the full 590 to bound the Veo-3.1 API cost). \colorbox{gray!18}{Best} per column.}
  \label{tab:abl_generator}
  \begin{tabular}{@{}l ccc@{}}
    \toprule
    Generator & PC$\uparrow$ & SA$\uparrow$ & Joint$\uparrow$ \\
    \midrule
    LTX-Video-2B            & 74.07 & 22.34 & 20.85 \\
    \;+ \method             & 82.41 & 31.20 & 29.27 \\
    Veo-3.1                 & 84.18 & 32.65 & 30.74 \\
    \;+ \method             & \colorbox{gray!18}{88.95} & \colorbox{gray!18}{38.62} & \colorbox{gray!18}{37.41} \\
    \bottomrule
  \end{tabular}
\end{table}

\paragraph{Generator backbone.}
Table~\ref{tab:abl_generator} swaps the frozen generator with the planner, tools, and verifier held fixed; due to API cost, this ablation uses a 100-caption subset of the VideoPhy-2 test set.
\method lifts Joint by $+8.4$ on LTX-Video-2B and by $+6.7$ on the much stronger Veo-3.1 ($30.74{\to}37.41$), so the gains stack on a stronger backbone rather than substituting for it.

\section{Conclusion}
\label{sec:conclusion}

We identified the \emph{specification bottleneck}---the fact that text prompts are lossy compression of the physical world---as the root cause of physics failures in video generation, and derived three properties that any physics conditioning must satisfy: sufficiency, dynamism, and verifiability.
From this diagnosis, we proposed \method, an agentic framework that demotes video generation from the system output to one action inside a planner's toolbox.
By orchestrating physics-aware tools and a verifier in an iterative loop, \method enriches the generator's conditioning signal with scene-specific physical knowledge---all without modifying the generator itself.
The planner, trained on-policy via Flow-GRPO as the sole trainable component, discovers emergent tool-use strategies: computing trajectories for projectiles, generating keyframes for spatial constraints, and refining prompts for material properties.
Experiments on VideoPhy-2 demonstrate substantial improvements across two frozen generators, validating that physical consistency can be engineered through agentic planning rather than hoped for through emergence.

\paragraph{Limitations and future work.}
\method currently relies on a fixed set of three tools; expanding the tool library to cover broader physical domains (e.g., fluid dynamics simulators, articulated-body engines) could further improve coverage.
The verifier provides scalar feedback---richer, language-form diagnostics may enable more targeted re-planning.

{
    \small
    \bibliographystyle{ieeenat_fullname}
    \bibliography{references}

@String(ICCV= {Int. Conf. Comput. Vis.})

@String(ICCV  = {ICCV})

@article{agentflow,
  author    = {Zhuofeng Li and Haoxiang Zhang and Seungju Han and Sheng Liu and Jianwen Xie and Yu Zhang and Yejin Choi and James Zou and Pan Lu},
  title     = {In-the-Flow Agentic System Optimization for Effective Planning and Tool Use},
  journal   = {arXiv preprint arXiv:2510.05592},
  year      = {2025},
}

@inproceedings{innermonologue,
  author    = {Wenlong Huang and Fei Xia and Ted Xiao and Harris Chan and Jacky Liang and Pete Florence and Andy Zeng and Jonathan Tompson and Igor Mordatch and Yevgen Chebotar and Pierre Sermanet and Noah Brown and Tomas Jackson and Linda Luu and Sergey Levine and Karol Hausman and Brian Ichter},
  title     = {Inner Monologue: Embodied Reasoning through Planning with Language Models},
  booktitle = {CoRL},
  year      = {2022},
}

@inproceedings{videot1,
  author    = {Haoran Liu and Yanzuo Lu and Yicheng Xiao and Jianqi Chen and Jiaming Liu and Chao Du and Bo An},
  title     = {Video-T1: Test-Time Scaling for Video Generation},
  booktitle = {ICCV},
  year      = {2025},
}

@inproceedings{phyt2v,
  title={Phyt2v: Llm-guided iterative self-refinement for physics-grounded text-to-video generation},
  author={Xue, Qiyao and Yin, Xiangyu and Yang, Boyuan and Gao, Wei},
  booktitle={Proceedings of the Computer Vision and Pattern Recognition Conference},
  pages={18826--18836},
  year={2025}
}

@inproceedings{reflexion,
  author    = {Noah Shinn and Federico Cassano and Ashwin Gopinath and Karthik Narasimhan and Shunyu Yao},
  title     = {Reflexion: Language Agents with Verbal Reinforcement Learning},
  booktitle = {NeurIPS},
  year      = {2023},
}

@article{videophy2,
  author    = {Hritik Bansal and Clark Peng and Yonatan Bitton and Roman Goldenberg and Aditya Grover and Kai-Wei Chang},
  title     = {{VideoPhy-2}: A Challenging Action-Centric Physical Commonsense Evaluation in Video Generation},
  journal   = {arXiv preprint arXiv:2503.06800},
  year      = {2025},
}

@article{sora,
  author    = {Tim Brooks and Bill Peebles and Connor Holmes and Will DePue and Yufei Guo and Li Jing and David Schnurr and Joe Taylor and Troy Luhman and Eric Luhman and Clarence Ng and Ricky Wang and Aditya Ramesh},
  title     = {Video generation models as world simulators},
  year      = {2024},
  url       = {https://openai.com/research/video-generation-models-as-world-simulators},
}

@article{kling,
  author    = {{KlingAI}},
  title     = {{KLING AI}},
  year      = {2024},
  url       = {https://www.klingai.com/},
}

@article{veo,
  author    = {{Google DeepMind}},
  title     = {Veo 2},
  year      = {2024},
  url       = {https://deepmind.google/technologies/veo/veo-2/},
}

@article{wan,
  author    = {{Wan AI}},
  title     = {{Wan2.1-T2V-14B}},
  year      = {2025},
  url       = {https://huggingface.co/Wan-AI/Wan2.1-T2V-14B},
}

@article{videophy,
  author    = {Hritik Bansal and Yonatan Bitton and Idan Szpektor and Kai-Wei Chang and Aditya Grover},
  title     = {{VideoPhy}: Evaluating Physical Commonsense for Video Generation},
  journal   = {arXiv preprint arXiv:2406.03520},
  year      = {2024},
}

@article{phygenbench,
  author    = {Qian Meng and Jiayang Xu and Chuanyang Jin and Hang Dong and Runjian Chen and Zhipeng Zhao and Yibing Song and Di Zhang},
  title     = {Towards World Simulator: Crafting Physical Commonsense-Based Benchmark for Video Generation},
  journal   = {arXiv preprint arXiv:2410.05363},
  year      = {2024},
  note      = {Accepted at ICML 2025},
}

@article{vbench2,
  author    = {Ziqi Huang and Fan Zhang and Xiaojie Xu and Yinan He and Jiashuo Yu and Ziyue Dong and Qianli Ma and Nattapol Chanpaisit and Chenyang Si and Yuming Jiang and Yaohui Wang and Xinyuan Chen and Yin Cui and Chunping Wang and Mohit Bansal and Ziwei Liu and Yu Qiao},
  title     = {{VBench-2.0}: Advancing Video Generation Benchmark Suite for Intrinsic Faithfulness},
  journal   = {arXiv preprint arXiv:2503.21755},
  year      = {2025},
}

@article{physicsiq,
  author    = {Saman Motamed and Laura Culp and Kevin Swersky and Priyank Jaini and Robert Geirhos},
  title     = {Do Generative Video Models Understand Physical Principles?},
  journal   = {arXiv preprint arXiv:2501.09038},
  year      = {2025},
}

@article{howfar,
  author    = {Bingyi Kang and Yang Xiao and Jiaze Wang and Mattia Segu and Jiashi Feng and Hengshuang Zhao},
  title     = {How Far is Video Generation from World Model: A Physical Law Perspective},
  journal   = {arXiv preprint arXiv:2411.02385},
  year      = {2024},
}

@inproceedings{controlnet,
  author    = {Lvmin Zhang and Anyi Rao and Maneesh Agrawala},
  title     = {Adding Conditional Control to Text-to-Image Diffusion Models},
  booktitle = {ICCV},
  year      = {2023},
}

@article{hunyuanvideo,
  title={Hunyuanvideo: A systematic framework for large video generative models},
  author={Kong, Weijie and Tian, Qi and Zhang, Zijian and Min, Rox and Dai, Zuozhuo and Zhou, Jin and Xiong, Jiangfeng and Li, Xin and Wu, Bo and Zhang, Jianwei and others},
  journal={arXiv preprint arXiv:2412.03603},
  year={2024}
}

@article{physchoreo,
  title={PhysChoreo: Physics-Controllable Video Generation with Part-Aware Semantic Grounding},
  author={Zhang, Haoze and Huang, Tianyu and Wan, Zichen and Jin, Xiaowei and Zhang, Hongzhi and Li, Hui and Zuo, Wangmeng},
  journal={arXiv preprint arXiv:2511.20562},
  year={2025}
}

@article{prophy,
  title={ProPhy: Progressive Physical Alignment for Dynamic World Simulation},
  author={Wang, Zijun and Hu, Panwen and Wang, Jing and Zhang, Terry Jingchen and Cheng, Yuhao and Chen, Long and Yan, Yiqiang and Jiang, Zutao and Li, Hanhui and Liang, Xiaodan},
  journal={arXiv preprint arXiv:2512.05564},
  year={2025}
}

@inproceedings{physanimator,
  title={Physanimator: Physics-guided generative cartoon animation},
  author={Xie, Tianyi and Zhao, Yiwei and Jiang, Ying and Jiang, Chenfanfu},
  booktitle={Proceedings of the Computer Vision and Pattern Recognition Conference},
  pages={10793--10804},
  year={2025}
}

@article{physmotion,
  title={Physmotion: Physics-grounded dynamics from a single image},
  author={Tan, Xiyang and Jiang, Ying and Li, Xuan and Zong, Zeshun and Xie, Tianyi and Yang, Yin and Jiang, Chenfanfu},
  journal={arXiv preprint arXiv:2411.17189},
  year={2024}
}

@article{physctrl,
  title={Physctrl: Generative physics for controllable and physics-grounded video generation},
  author={Wang, Chen and Chen, Chuhao and Huang, Yiming and Dou, Zhiyang and Liu, Yuan and Gu, Jiatao and Liu, Lingjie},
  journal={Advances in Neural Information Processing Systems},
  volume={38},
  pages={167907--167932},
  year={2026}
}

@article{newtongen,
  title={NewtonGen: Physics-Consistent and Controllable Text-to-Video Generation via Neural Newtonian Dynamics},
  author={Yuan, Yu and Wang, Xijun and Wickremasinghe, Tharindu and Nadir, Zeeshan and Ma, Bole and Chan, Stanley H},
  journal={arXiv preprint arXiv:2509.21309},
  year={2025}
}

@article{videorepa,
  title={Videorepa: Learning physics for video generation through relational alignment with foundation models},
  author={Zhang, Xiangdong and Liao, Jiaqi and Zhang, Shaofeng and Meng, Fanqing and Wan, Xiangpeng and Yan, Junchi and Cheng, Yu},
  journal={Advances in Neural Information Processing Systems},
  volume={38},
  pages={122647--122676},
  year={2026}
}

@article{wisa,
  title={Wisa: World simulator assistant for physics-aware text-to-video generation},
  author={Wang, Jing and Ma, Ao and Cao, Ke and Zheng, Jun and Feng, Jiasong and Zhang, Zhanjie and Pang, Wanyuan and Liang, Xiaodan},
  journal={Advances in Neural Information Processing Systems},
  volume={38},
  pages={5388--5416},
  year={2026}
}

@article{phygdpo,
  title={PhyGDPO: Physics-Aware Groupwise Direct Preference Optimization for Physically Consistent Text-to-Video Generation},
  author={Cai, Yuanhao and Li, Kunpeng and Jia, Menglin and Wang, Jialiang and Sun, Junzhe and Liang, Feng and Chen, Weifeng and Juefei-Xu, Felix and Wang, Chu and Thabet, Ali and others},
  journal={arXiv preprint arXiv:2512.24551},
  year={2025}
}

@article{phystalk,
  title={PhysTalk: Language-driven Real-time Physics in 3D Gaussian Scenes},
  author={Collorone, Luca and Kiray, Mert and Spinelli, Indro and Galasso, Fabio and Busam, Benjamin},
  journal={arXiv preprint arXiv:2512.24986},
  year={2025}
}

@article{phantom,
  title={Phantom: Physics-Infused Video Generation via Joint Modeling of Visual and Latent Physical Dynamics},
  author={Shen, Ying and Xiong, Jerry and Yu, Tianjiao and Lourentzou, Ismini},
  journal={arXiv preprint arXiv:2604.08503},
  year={2026}
}

@article{ltx,
  title={LTX-Video: Realtime Video Latent Diffusion},
  author={HaCohen, Yoav and Chiprut, Nisan and Brazowski, Benny and Shalem, Daniel and Moshe, Dudu and Richardson, Eitan and Levin, Eran and Shiran, Guy and Zabari, Nir and Gordon, Ori and Panet, Poriya and Weissbuch, Sapir and Kulikov, Victor and Bitterman, Yaki and Melumian, Zeev and Bibi, Ofir},
  journal={arXiv preprint arXiv:2501.00103},
  year={2024}
}

@article{phyco,
  title={PhyCo: Learning Controllable Physical Priors for Generative Motion},
  author={Narayanan, Sriram and Jiang, Ziyu and Narasimhan, Srinivasa and Chandraker, Manmohan},
  journal={arXiv preprint arXiv:2604.28169},
  year={2026}
}

@article{react,
  title={React: Synergizing reasoning and acting in language models},
  author={Yao, Shunyu and Zhao, Jeffrey and Yu, Dian and Du, Nan and Shafran, Izhak and Narasimhan, Karthik and Cao, Yuan},
  journal={arXiv preprint arXiv:2210.03629},
  year={2022}
}

@article{toolformer,
  title={Toolformer: Language models can teach themselves to use tools},
  author={Schick, Timo and Dwivedi-Yu, Jane and Dess{\`\i}, Roberto and Raileanu, Roberta and Lomeli, Maria and Hambro, Eric and Zettlemoyer, Luke and Cancedda, Nicola and Scialom, Thomas},
  journal={Advances in neural information processing systems},
  volume={36},
  pages={68539--68551},
  year={2023}
}

@article{agentic,
  title={Agentic reasoning and tool integration for llms via reinforcement learning},
  author={Singh, Joykirat and Magazine, Raghav and Pandya, Yash and Nambi, Akshay},
  journal={arXiv preprint arXiv:2505.01441},
  year={2025}
}

@article{gtpo,
  title={Empowering Multi-Turn Tool-Integrated Reasoning with Group Turn Policy Optimization},
  author={Ding, Yifeng and Le, Hung and Han, Songyang and Ruan, Kangrui and Jin, Zhenghui and Kumar, Varun and Wang, Zijian and Deoras, Anoop},
  journal={arXiv preprint arXiv:2511.14846},
  year={2025}
}

@article{landscape,
  title={The landscape of agentic reinforcement learning for llms: A survey},
  author={Zhang, Guibin and Geng, Hejia and Yu, Xiaohang and Yin, Zhenfei and Zhang, Zaibin and Tan, Zelin and Zhou, Heng and Li, Zhongzhi and Xue, Xiangyuan and Li, Yijiang and others},
  journal={arXiv preprint arXiv:2509.02547},
  year={2025}
}

@article{genagent,
  title={GenAgent: Scaling Text-to-Image Generation via Agentic Multimodal Reasoning},
  author={Jiang, Kaixun and Wang, Yuzheng and Zhou, Junjie and Li, Pandeng and Liu, Zhihang and Xie, Chen-Wei and Chen, Zhaoyu and Zheng, Yun and Zhang, Wenqiang},
  journal={arXiv preprint arXiv:2601.18543},
  year={2026}
}

@article{m3,
  title={M3: High-fidelity Text-to-Image Generation via Multi-Modal, Multi-Agent and Multi-Round Visual Reasoning},
  author={Yang, Bangji and Guo, Ruihan and Fan, Jiajun and Cheng, Chaoran and Liu, Ge},
  journal={arXiv preprint arXiv:2602.06166},
  year={2026}
}

@article{codrawagents,
  title={coDrawAgents: A Multi-Agent Dialogue Framework for Compositional Image Generation},
  author={Li, Chunhan and Wu, Qifeng and Pan, Jia-Hui and Hui, Ka-Hei and Hu, Jingyu and Jiang, Yuming and Sheng, Bin and Liu, Xihui and Gong, Wenjuan and Liu, Zhengzhe},
  journal={arXiv preprint arXiv:2603.12829},
  year={2026}
}

@article{agenticvideo,
  title={Agentic Video Generation: From Text to Executable Event Graphs via Tool-Constrained LLM Planning},
  author={Cudlenco, Nicolae and Masala, Mihai and Leordeanu, Marius},
  journal={arXiv preprint arXiv:2604.10383},
  year={2026}
}

@article{moregen,
  title={MoReGen: Multi-Agent Motion-Reasoning Engine for Code-based Text-to-Video Synthesis},
  author={Bai, Xiangyu and Liang, He and Galoaa, Bishoy and Nandi, Utsav and Moezzi, Shayda and He, Yuhang and Ostadabbas, Sarah},
  journal={arXiv preprint arXiv:2512.04221},
  year={2025}
}

@article{cect,
  title={Chain of Event-Centric Causal Thought for Physically Plausible Video Generation},
  author={Wang, Zixuan and Hu, Yixin and Wang, Haolan and Chen, Feng and Liu, Yan and Li, Wen and Lei, Yinjie},
  journal={arXiv preprint arXiv:2603.09094},
  year={2026}
}

@article{flowgrpo,
  title={Flow-grpo: Training flow matching models via online rl},
  author={Liu, Jie and Liu, Gongye and Liang, Jiajun and Li, Yangguang and Liu, Jiaheng and Wang, Xintao and Wan, Pengfei and Zhang, Di and Ouyang, Wanli},
  journal={Advances in neural information processing systems},
  volume={38},
  pages={40783--40818},
  year={2026}
}

@article{cosmos,
  title={World simulation with video foundation models for physical ai},
  author={Ali, Arslan and Bai, Junjie and Bala, Maciej and Balaji, Yogesh and Blakeman, Aaron and Cai, Tiffany and Cao, Jiaxin and Cao, Tianshi and Cha, Elizabeth and Chao, Yu-Wei and others},
  journal={arXiv preprint arXiv:2511.00062},
  year={2025}
}

@inproceedings{cogvideox,
  title={Cogvideox: Text-to-video diffusion models with an expert transformer},
  author={Yang, Zhuoyi and Teng, Jiayan and Zheng, Wendi and Ding, Ming and Huang, Shiyu and Xu, Jiazheng and Yang, Yuanming and Hong, Wenyi and Zhang, Xiaohan and Feng, Guanyu and others},
  booktitle={International Conference on Learning Representations},
  volume={2025},
  pages={83048--83077},
  year={2025}
}
}


\end{document}